\documentclass[letterpaper, 10 pt, conference]{ieeeconf}  % Comment this line out if you need a4paper

\IEEEoverridecommandlockouts                              % This command is only needed if 
\usepackage{booktabs} % For formal tables
\usepackage{graphicx, caption, subcaption}
\usepackage{flushend}

\usepackage{hyperref} 
\usepackage{amsmath}
\usepackage{algorithm}
\usepackage[noend]{algpseudocode}
\usepackage{siunitx}
\usepackage{graphbox}
\usepackage{xcolor}

\usepackage{amsmath}

\usepackage[noadjust]{cite}

\title{\LARGE \bf
Sim-to-Real Learning for Casualty Detection \\ 
from Ground Projected Point Cloud Data
}

\author{Roni Permana Saputra$^{1,2}$, Nemanja Rakicevic$^{1}$, Petar Kormushev$^{1}$% <-this % stops a space
\thanks{$^{1}$Authors are with Robot Intelligence Lab, Dyson School of Design Engineering, Imperial College London, UK
        {\tt\small \{r.saputra, n.rakicevic, p.kormushev\}@imperial.ac.uk}}%
\thanks{$^{2}$Roni P. Saputra is also with the Research Center for Electrical Power and Mechatronics, Indonesian Institute of Sciences - LIPI, Indonesia}%
}

\begin{document} 

\maketitle
\thispagestyle{empty}
\pagestyle{empty}

%%%%%%%%%%%%%%%%%%%%%%%%%%%%%%%%%%%%%%%%%%%%%%%%%%%%%%%%%%%%%%%%%%%%%%%%%%%%%%%%
\begin{abstract}
This paper addresses the problem of human body detection---particularly a human body lying on the ground (a.k.a. casualty)---using point cloud data. 
This ability to detect a casualty is one of the most important features of mobile rescue robots, in order for them to be able to operate autonomously. 
We propose a deep-learning-based casualty detection method using a deep convolutional neural network (CNN). 
This network is trained to be able to detect a casualty using a point-cloud data input. 
In the method we propose, the point cloud input is pre-processed to generate a depth image-like ground-projected heightmap. 
This heightmap is generated based on the projected distance of each point onto the detected ground plane within the point cloud data. 
The generated heightmap---in image form---is then used as an input for the CNN to detect a human body lying on the ground. 
To train the neural network, we propose a novel sim-to-real approach, in which the network model is trained using synthetic data obtained in simulation and then tested on real sensor data. 
To make the model transferable to real data implementations, during the training we adopt specific data augmentation strategies with the synthetic training data. 
The experimental results show that data augmentation introduced during the training process is essential for improving the performance of the trained model on real data.
More specifically, the results demonstrate that the data augmentations on raw point-cloud data
have contributed to a considerable improvement of the trained model performance.
\end{abstract}

%%%%%%%%%%%%%%%%%%%%%%%%%%%%%%%%%%%%%%%%%%%%%%%%%%%%%%%%%%%%%%%%%%%%%%%%%%%%%%%%

\section{INTRODUCTION}
Detecting injured people, i.e. casualties, during search and rescue (SAR) missions is a key challenge faced in the area of SAR robotics. 
%
% To achieve fast and accurate responses during a mission, autonomous casualty detection using a SAR robot is essential.
%
To be able to detect a casualty, a SAR robot needs to perceive one or more physical properties of the casualty, such as visual features, temperature, scent and 3D body shape.
% ~\cite{de2009human,burion2004human}.
%[DO REFERENCES 1, 2 USE ALL THE MENTIONED PROPERTIES (visual features, temperature, scent)? IF NOT ADD THE REFERENCES]
% Answer : YES
%
A wide range of research studies have been conducted relating to searching and detecting human bodies and leveraging the advances in development for various sensors that can perceive these physical properties~\cite{ivanovs2019multisensor,andriluka2010vision}. 
%[REFERENCES? WHICH STUDIES] 

% Since RGB-D  cameras that can provide both visual information and 3D point cloud data are now widely available and affordable, this has attracted the development of various object detection techniques---including human body detection---that make use of this sensor~\cite{Gupta2014LearningRF,Song2016DeepSS}. 
% %
% A number of successful research studies have been conducted to develop methods of human presence detection~\cite{teixeira2010survey,ogale2006survey}. 
% %
% A number of these have looked into detecting human presence using 3D point cloud data and have reported successful implementations~\cite{aggarwal2014human,chen2013survey}. 
% %

%
Recently, significant progress has been made in developing human presence detection techniques for general cases, such as pedestrian detection and human activity capturing~\cite{dollar2012pedestrian,hu2004survey}. 
%
% Most of the techniques deployed work best for detecting people who are fully visible and presented in certain pose variations, such as standing, walking and sitting. 
Most of these techniques perform optimally when detecting people who are fully visible and presented in certain pose variations, such as standing, walking and sitting.
Moreover, in most cases, the sensors or cameras are used to detect the person’s face perpendicularly with respect to the object of interest, because otherwise additional image processing is needed to align them properly~\cite{Cao2012FaceAB}.
%DONE
%[CITE \url{https://www.microsoft.com/en-us/research/wp-content/uploads/2013/01/Face-Alignment-by-Explicit-Shape-Regression.pdf}]. 
%

The following particular cases are still under-explored problems: 

\begin{figure}[t]
\centering
\includegraphics[width=0.9\linewidth]{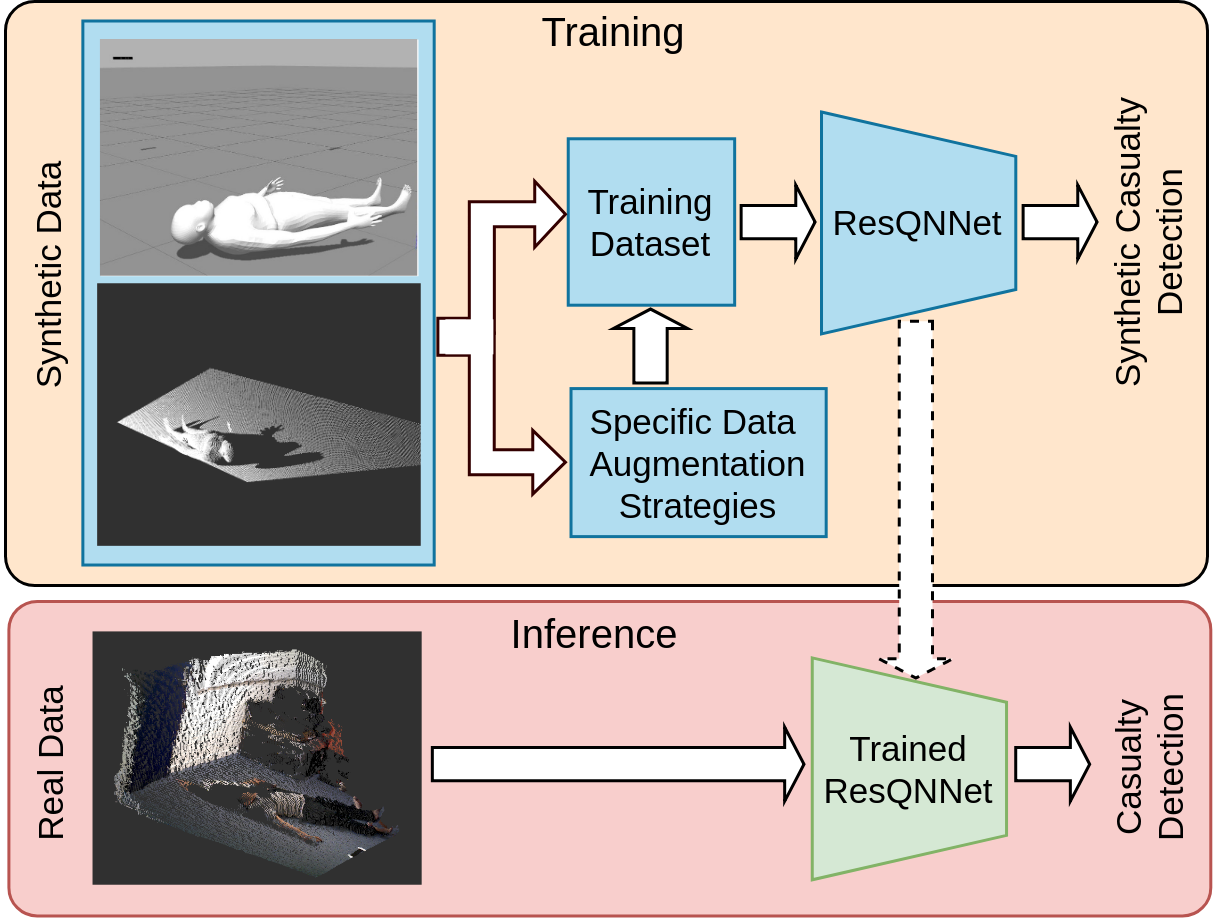}
\caption{Sim-to-real casualty detection learning pipeline.\\ 
\textbf{Top (learning stage)}: The proposed ResQNNet framework is trained using synthetic (augmented) data obtained from simulation; 
\textbf{Bottom (inference stage)}: The trained ResQNNet uses real sensor data to perform casualty detection.}
\label{fig:synth-to-real}
\vspace{-1.3em}
\end{figure}

\begin{enumerate}
\item Detecting injured people, usually a human body lying on the ground. This case leads to significant difference in pose variation compared to with standard human presence detection techniques. Moreover, the fact that the floor is right under the body makes depth segregation extremely challenging; 

\item Using an onboard sensor or camera on ground robots (facing forward) to detect a human body lying on the ground. This scenario introduces a limited viewing angle and permits the camera to only partially observe the object of interest.

\end{enumerate}

On the other hand, the data-driven approach, such as deep convolutional neural networks (CNNs), is currently popular in various research areas. 
Particularly for visual perception tasks, leveraging various advanced techniques using CNNs has achieved considerable progress~\cite{szegedy2015going,girshick2014rich,girshick2015fast,ren2017faster,lin2018focal,he2017mask,peng2018megdet}.
However, to achieve such good performance, it is well known that this data-driven approach requires an extensive amount of data. 
In practice, a large amount of training data is not always available, especially for cases that are problem specific. 
%
% Generating new datasets is also not trivial work and is expensive. 
Also, generating such new datasets is expensive and non-trivial work.
To the best of our knowledge, there is currently no available dataset for detecting a human body lying on the floor using sensors onboard a mobile ground robot.
% using onboard sensors on a ground mobile robot. 

In this study, we propose a data-driven approach for detecting a casualty using a 3D point cloud data input. 
We propose a sim-to-real learning technique (see Fig.~\ref{fig:synth-to-real}), in which we use a synthetic dataset---generated from simulations---to train a deep CNN and then apply the trained model to real sensor data. 
To achieve a good performance in inferring from real data, we introduce specific data augmentation strategies, such as noise augmentation, down-sampling and segment removal, for augmenting the simulated training data. 
The main contributions of this work include: 
\begin{enumerate}
\item Introducing ResQNNet, a novel framework for deep-learning-based casualty detection using a point-cloud input;
\item Exploring specific data augmentation strategies to improve the sim-to-real performance of the proposed casualty detection approach;
\item Making publicly available our novel dataset consisting of synthetic and real 3D point-cloud data, containing human bodies lying on the ground upon publication.~\footnote{https://sites.google.com/view/sim-to-real-resqnnet/}
\end{enumerate}

\section{RELATED WORK}

Rapid developments in advanced machine learning, especially in the area of deep learning techniques---such as deep CNNs---has increased the use of these techniques in a wide range of computer vision research areas~\cite{szegedy2015going,girshick2015fast,girshick2014rich,he2017mask,peng2018megdet,lin2018focal,ren2017faster}. 
These CNN-based techniques rely significantly on a large amount of training data to achieve an acceptable performance and generalisation. 
However, in many cases, such large datasets are not always openly available. 
For instance, to the best of our knowledge, currently, there is no available existing dataset that is suitable for the particular purpose of our study, which utilises point cloud data inputs to detect a human body lying on the ground. 
Generating datasets from real sensor data is not trivial and manually labelling this dataset is prohibitively expensive as well. 

% Learning from synthetic data
Work focusing on learning tasks from synthetic images has been extensively studied in recent years~\cite{marin2010learning,papon2015semantic,varol2017learning,pishchulin2012articulated,chen2016synthesizing,rahmani2015learning,rahmani20163d,pishchulin2011learning}. 
For instance, in~\cite{pishchulin2012articulated}, the researchers present studies on learning from synthetic images for 2D human pose estimation tasks. 
Tasks involving 3D pose estimations learned from synthetic data is also explored in~\cite{chen2016synthesizing}, and in~\cite{rahmani2015learning,rahmani20163d} the authors investigate the subject by presenting action recognition tasks using a similar sim-to-real technique. 
Particularly in the task of human presence detection, in the works presented in~\cite{marin2010learning,pishchulin2011learning}, learning from synthetic data is used to complete pedestrian detection tasks. 
In more recent work on domain adaptation~\cite{Bousmalis2017UnsupervisedPD}, the authors use a mixture of synthetic, real and realistic generated images to extract invariant features that enable a robot to learn a policy in simulation which would successfully transfer on a real robot. Alternatively, in the domain randomisation approach~\cite{Tobin2017DomainRF}, a robust policy is learned by exposing the learner to a large number of various simulated scenarios which might encompass cases similar to the real ones.

To increase the resemblance between the synthetic image and the real image and increase the learning performance test with the real image, several post-processing scenarios can be injected onto the synthetic image. 
Rozantsev et al~\cite{rozantsev2015rendering} describe their study on the technique of rendering synthetic images in their post-processing scenarios. 
These post-processing scenarios include: object boundary blurring (BB), motion blurring (MB), random noise (RN) and material properties (MPs).
Planche et al in~\cite{planche2017} generate synthetic depth data from 3D models. 
In this work, they try to understand the causes of the noise in the real depth data and mimic this noise in synthetic depth data. 
Similarly, in the work in~\cite{rozantsev2015rendering}, RN and MB are added to the synthetic data. 
In addition to these noises, they also include radial and tangential lens distortion, lens scratching and lens graining into the data. %

%[TRY TO WRITE IN ONE SENTENCE WHY THIS WORK IS DIFFERENT FROM THE OTHERS?]

\section{ROBOT PLATFORM AND HARDWARE}

\begin{figure}[t]
\centering
\includegraphics[width=0.9\linewidth]{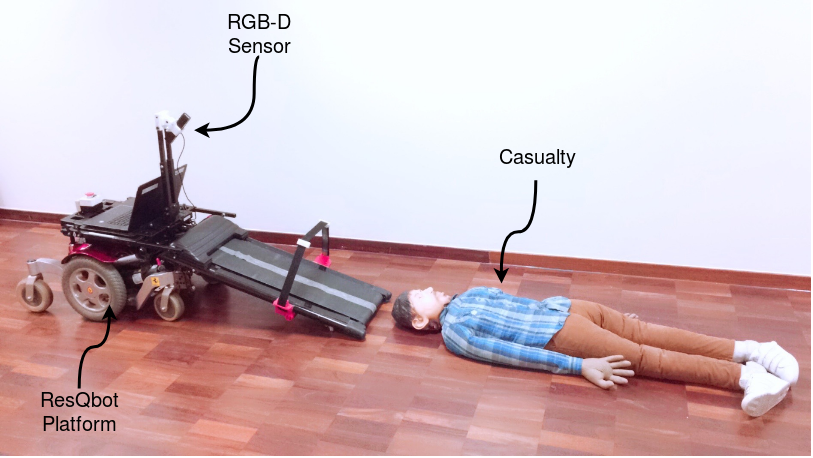}
% \vspace{-0.2cm}
\caption{ResQbot platform equipped with an RGB-D camera for detecting a casualty~\cite{Saputra2018ResQbotAM,Saputra2018ResQbotTaros}.}
\label{fig:resqbot}
\vspace{-0.5cm}
\end{figure}

\begin{figure*}[!t]
\centering
\includegraphics[width=5.4in]{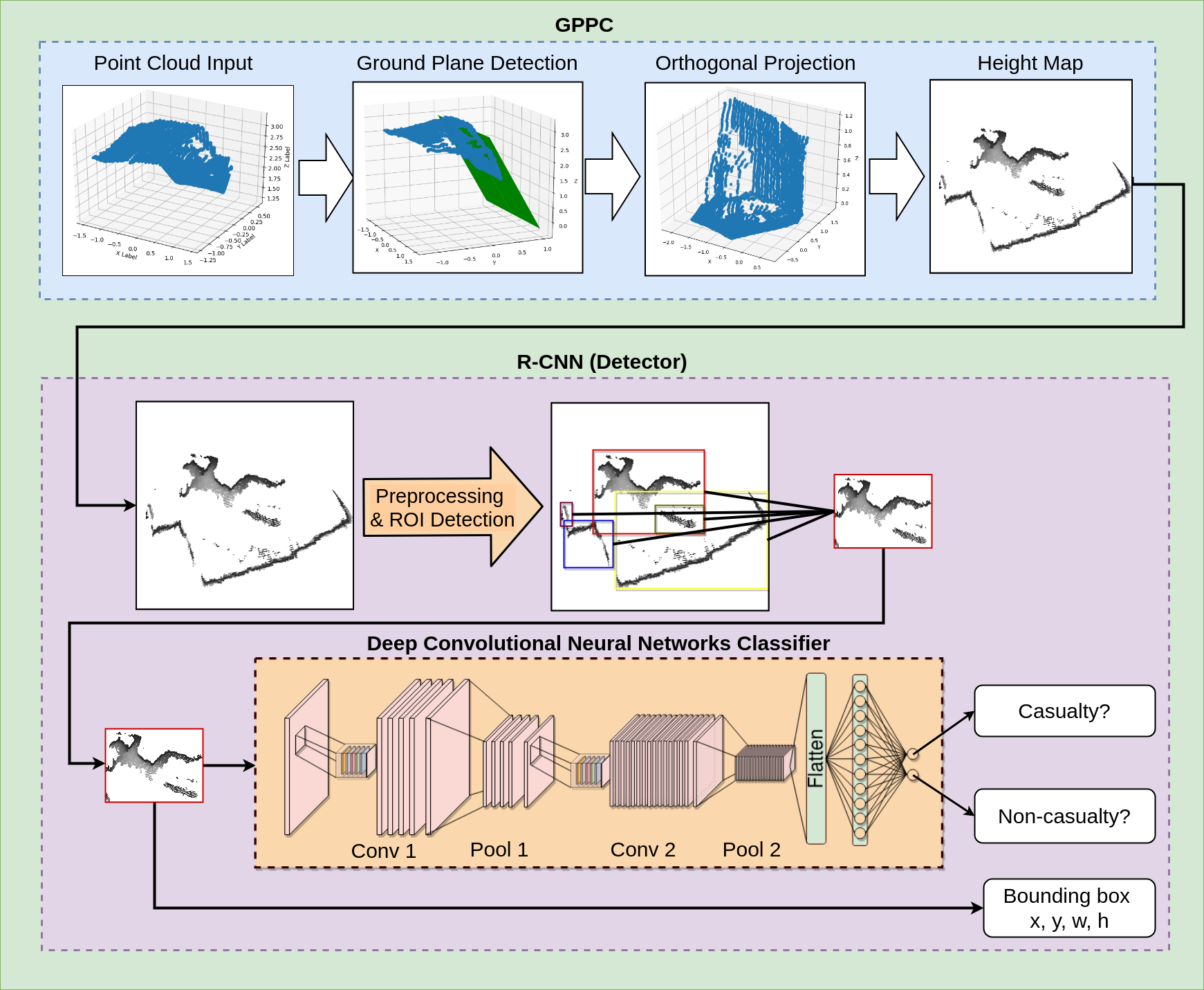}
\caption{ResQNNet - proposed deep-learning-based casualty detection framework. It consists of the ground projected point cloud (GPPC) part (above) and the object detection and classification part (below).}
\label{fig:resqnnet}
\vspace{-1.5em}
\end{figure*}

In previous work~\cite{Saputra2018ResQbotAM,Saputra2018ResQbotTaros}, we have developed and tested a mobile rescue robot called ResQbot (as shown in Fig. 2). 
This robot is designed so that it can safely perform a casualty extraction task with human victims in rescue scenarios. 
To perform this casualty extraction procedure, a loco-manipulation approach is used, in which the robot uses its locomotion system---wheeled locomotion---to perform a manipulation task.
An example of such a task used in SAR missions is loading the victim from the ground. 

ResQbot is equipped with perception devices, including an RGB-D camera, that provides the perceptive feedback required during the operation. 
The RGB-D camera is also designed to further enable ResQbot to perform autonomous casualty detection. 
In this study, in particular, we used real data---i.e. point cloud data---obtained from the ResQbot sensor as part of our experiments. 
The work conducted in this study is also designed so that it can be implemented in real-time as part of the ResQbot system in future work.  
We refer readers to~\cite{Saputra2018ResQbotAM,Saputra2018ResQbotTaros} for more details about the ResQbot design and specification. 

% ###################################################################
\section{METHODOLOGY}

\subsection{Problem Definition}
The problem of casualty detection addressed in this study %is associated with the problem of human body detection, 
concerns when the body is lying on the ground. 
In contrast, most state-of-the-art human detection studies usually concern human bodies in upright poses. 
%[THIS IS NOT COMPLETELY TRUE, LOOK AT THIS \url{https://arxiv.org/pdf/1802.00434.pdf} and \url{https://arxiv.org/pdf/1611.08050.pdf} BUT YOU CAN SAY AGAIN THAT THERE IS NOT ENOUGH DATA FOR CASUALTY POSE HUMAN DETECTION. ALSO SAY HERE THAT THEY USE IMAGES BUT YOU NEED POINT CLOUD AND EXPLAIN WHY...SPECIFIC TO SEARCH AND RESCUE BLABLABLA]
%
Moreover, in most cases the image is captured by the camera being faced perpendicularly towards the object (i.e. the person). 

In our study we aim to address a number of problems, which include: 

\begin{enumerate}
\item Detecting a human body lying on the ground;
\item Using an onboard sensor on a mobile robot that is not facing perpendicularly towards the ground and the body;
\item Using a 3D point cloud as the data input.
\end{enumerate}

To deal with this problem, we propose a data-driven approach based on training a deep neural network to detect a human body lying on the floor using a ground projected point cloud (GPPC) input. 
We introduce a novel framework called ResQNNet for this deep-learning-based casualty detection from point cloud data.
To ensure this data-driven approach performs well, it is essential to have a substantial amount of training data.

To the best of our knowledge, currently, there is no dataset available for training neural networks for our particular setup (i.e. input and output data types). 
However, collecting and labelling a real dataset using real sensor data is very expensive. 
Therefore, in this study, we propose to use synthetic data for training purposes. 
This data is generated using Gazebo simulations. 
In each simulation we use a virtual RGB-D sensor that produces point-cloud data and loads human body models into the simulation, in different positions and orientations with respect to the sensors. 
% We generated 216 models which vary in size, gender and age.

To ensure that the ResQNNet trained on synthetic data is transferable to real sensor data, in addition to the standard data augmentation usually implemented in deep learning we propose additional specific augmentation strategies presented in section V.D. 
%[WHICH ARE THE USUAL ONES???] 
% Fig. 1 illustrates the general pipeline of the experiments in this study. 

\subsection{ResQNNet Architecture}

In this section, we present our proposed approach to the casualty detection problem defined in section IV.A. The full architecture of ResQNNet is presented in Fig. 3.
The architecture is based on integrating the proposed GPPC casualty detection technique presented by Saputra et al.~\cite{Saputra2018CasualtyDF,Saputra2018} and adopting a region-based convolutional network (R-CNN) for object detection~\cite{girshick2014rich}. 
The classification part is adopted from the LeNet architecture, where we refine several hyperparameters for our own particular purposes (i.e. the input and output structure).
%
% We choose LeNet architecture for classification part in this study since this is the very beginning deep CNN architecture for simple classification task.
%
The LeNet architecture is simple but expressive enough for our classification task.
Our hypothesis is that the proposed method can still produce good results on detecting casualties, even though we are using a basic CNN architecture.

\subsubsection{Generating GPPC heightmap}

The initial step in the proposed approach is generating a heightmap from the point cloud data input projected onto the ground plane. In order to generate this heightmap, first we estimate a ground plane from the point cloud data input. We refer readers to~\cite{Saputra2018} for more details about this GPPC approach.
Contrary to the previous work in~\cite{Saputra2018}, here we generate a heightmap representation from the projected points instead of a discrete grid-map to preserve 3D information of the objects, including casualties.
The heightmap generated from GPPC is in the form of greyscale grid cells representing the maximum normalised distance of point pairs occupying each cell.  

Let $\mathbf{C}$ represent an $m\times m$ 2D grid cells on the ground plane. 
Then find the point pair that corresponds to each cell component.
If there is at least one point pair that corresponds to the cell, the greyscale value of this cell is the maximum normalised distance of  the point pairs occupying the cell.
On the other hand, if there are no point pairs occupying the cell, we assign the maximum greyscale value to this cell.
%
% The value of each grid cell can be obtained using:
% %
% % Fig. --- shows an example of projected point cloud (i.e. GPPC) and the corresponding generated heightmap.
% %[HOW CAN YOU CALCULATE A NORM OF A DISTANCE? IS DISTANCE A SCALAR OR A VECTOR?]
% \begin{equation*}
%     \mathbf{C}(i)=\left\{\begin{matrix}
%     \max (\frac{d_j^i}{D}) & \textup{if} \; \; \mathbf{C}(i)\; \; \textup{is occupied}\\ 
%     1 & \textup{otherwise.}
%     \end{matrix}\right.
%   \label{eq:heightmap}
% % \vspace{-0.22cm}
% \end{equation*}

\subsubsection{Region of interest (ROI) based on contour detection}
The next part of the ResQNNet architecture is detecting the ROI from the GPPC heightmap. 
Regarding the ROI detection presented in the R-CNN paper, a variety of sophisticated methods have been proposed for generating region proposals from the RGB image. 
However, in our case in particular, the heightmap produced using the GPPC process consists of greyscale images where the majority of the pixels not belonging to objects are white. 
Therefore, we hypothesise that simple contour detection is sufficient to generate a region proposal in this case. 
We use the OpenCV library to find contours through discretising an image.  
To minimise the number of region proposals, we filter the detected contours within the image. We select the detected contours to be considered as candidate objects of interest, based on the contour size being above a certain threshold.
%(see Equation~\ref{eq:contour-filter}) [IS THIS EQUATION NECESSARRY? IT quite obvious]

%\begin{equation}
%  ROI(i)=\begin{cases}
%    True, & \text{if $area(i) > threshold$}\\
%    False, & \text{otherwise}
%  \end{cases}
%    \label{eq:contour-filter}
%\end{equation}

\subsubsection{CNN for classifying the ROIs}

The final part of ResQNNet is the CNN classifier. The particular CNN architecture used in ResQNNet is adopted from the LeNet architecture presented in~\cite{LeCun2006PROCOT}. 
The networks consist of feature extraction parts with two layers of convolutions with max-pooling and downsampling at each output of the convolution layer. The output of the feature extraction part is then flattened and passed to a fully connected layer and softmax classifier layer. 
The output of this CNN is a binary classifier, that classifies each candidate region passed to the network into two classes, casualty and non-casualty. 
%

% %
% We have experimented with the hyper-parameters of the CNN network, including the kernel size, the filter number and the pooling parameter.
% %
% We aim to find the most optimal and simplest structure that can sufficiently achieve good results for our classification task.
% % %
% From our experimental study, the parameter setup listed in the table 1. demonstrate optimal performance for our particular tasks associated with experiments presented in this paper.

% \begin{table}[b]
% \begin{center}
%   \caption{Parameter setup of ResQNet}
%   \label{tab:comparison}
%   \begin{tabular}{lccccl}
%     \toprule
%     Layer&Input Size&Kernel Size&Stride &Pad&Output Size\\
%     \midrule

%   \bottomrule
% \end{tabular}
% \end{center}
% \end{table}

\section{OBTAINING THE DATASET}

% As it illustrated in Fig. 1, our pipeline in this study consists of training ResQNet using large amount of synthetic data and testing the trained network using real sensor data.
% %
% This section describe our setup for obtaining dataset for both training, validation and testing of ResQNet.
% %

\subsection{Generating Synthetic Human Body Models from an On-line Human Body Shape Modeller }

The synthetic human bodies used in our experiments are created via an on-line human body modeller developed by the University of Michigan Transportation Research Institute (UMTRI)~\cite{Reed2014DevelopingAI}. Fig. 4 illustrates the four different human body shapes generated by a set of different parameters in the on-line modellers~\cite{WinNT}. For the purpose of generating training datasets, we create 216 synthetic human bodies by combining different value settings, including the stature, body mass index (BMI), the ratio of erect sitting height to stature (SHS)  and age. 
% DONE
%[IT IS OK TO REFER TO 41 BUT DO NOT PUT UNKNOWN ABBREVIATIONS, WRITE WHAT THEY MEAN AND WHAT ARE THEIR UNITS. ADD UNITS TO TABLE 1]
%
We refer readers to~\cite{Reed2014DevelopingAI} for more details about this on-line modeller and the parameter setting for this modeller. 
%
% \textcolor{red}{In addition to this, we also vary the orientation of the body w.r.t. the camera.}
In addition to this, we also vary the orientation of the body w.r.t. the camera.
Table I shows the combination of value settings that we use for generating the synthetic bodies. 

\begin{table}[t]
\begin{center}
  \caption{The combination of value settings for generating the synthetic bodies.}
  \label{tab:body-parameter}
  \begin{tabular}{rc}
    \toprule
    Parameter&Value Range\\
    \midrule
    Stature [cm] &1500, 1600, 1700, 1800, 1900, 2000 \\
    BMI [$kg/m^2$] &20, 25, 30 \\
    SHS [ratio] &0.4, 0.5, 0.6 \\
    Age [years]&20, 40, 60, 80 \\
    Orientation [deg]&0, $\pm$ 45, $\pm$ 90, $\pm$ 135, 180\\
    \bottomrule
  \end{tabular}
\end{center}
\vspace{-0.2cm}
\end{table}

\begin{figure}[t]
\centering
\includegraphics[width=2.8in]{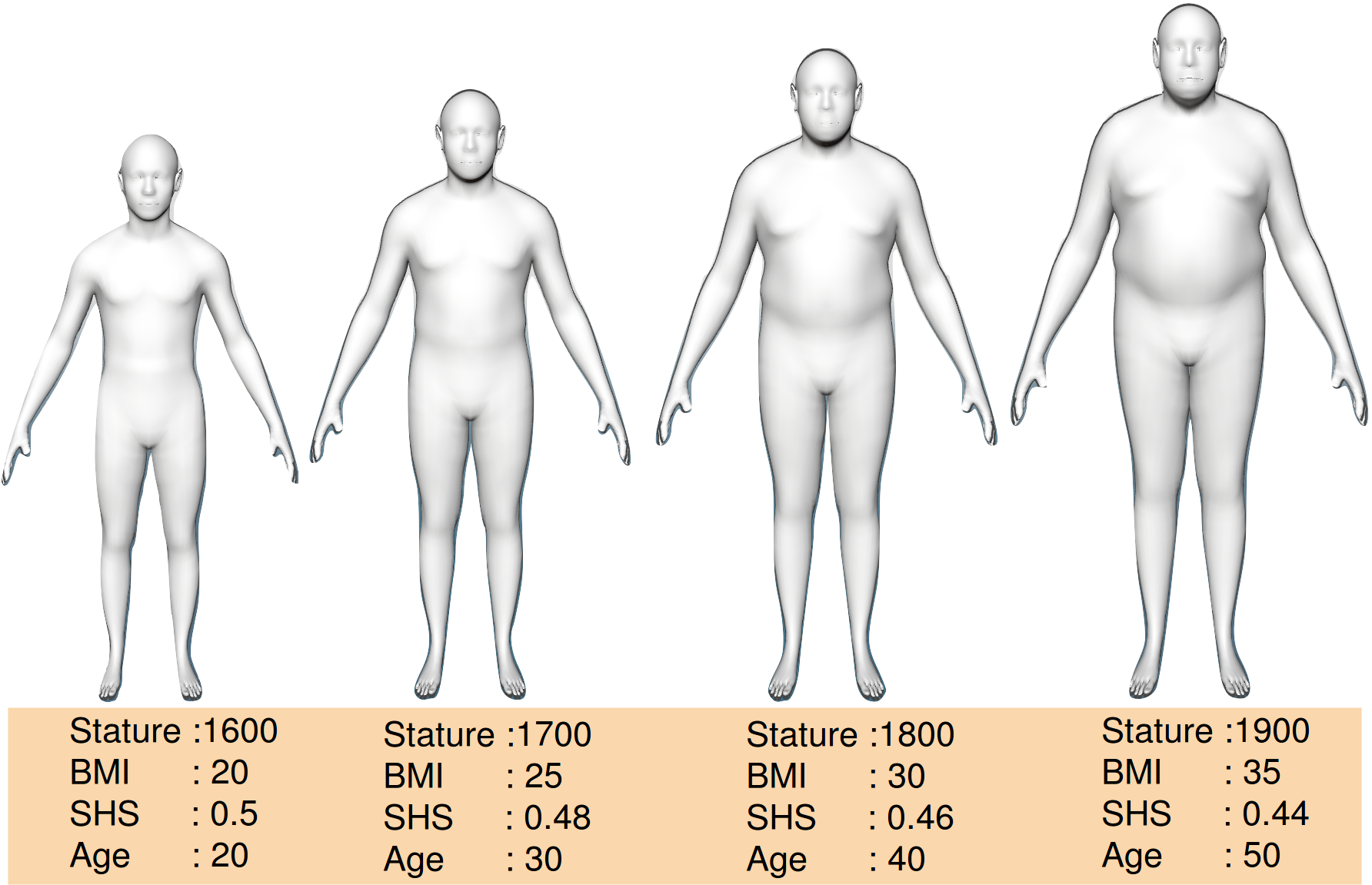}
\caption{Various human body models created via on-line human body modeller from UMTRI.}
\label{fig:body-model}
\vspace{-0.5cm}
\end{figure}

\subsection{Generating Simulated Point Cloud Data Using Gazebo Simulation}
\begin{figure}[t]
\centering
\includegraphics[width=3.1in]{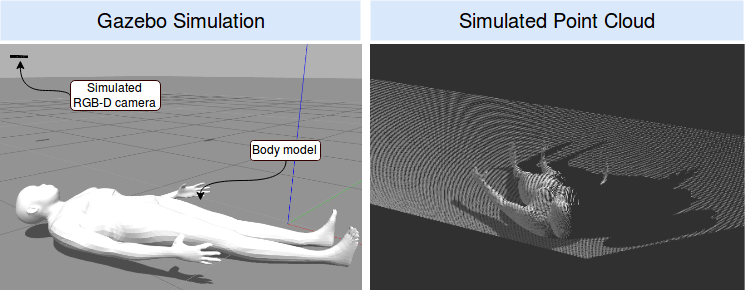}
\caption{Obtaining synthetic data from the Gazebo simulator.\\
\textbf{Left}: Loading human body model into Gazebo and simulating a virtual RGB-D sensor. \textbf{Right}: Visualisation of synthetic point-cloud data generated within the Gazebo simulator.}
\label{fig:body-Gazebo}
\vspace{-0.2em}
\end{figure}

To obtain synthetic point cloud data, we use Gazebo simulation. In this simulation, we simulate an RGB-D camera that can produce point cloud data simulation. The human body models created are then loaded into the Gazebo and set into various positions and orientations with respect to the RGB-D camera.
In total we generate 10 thousand synthetic point cloud data with synthetic human body inside and another 10 thousand synthetic point cloud data with no synthetic human body inside.
Fig.~\ref{fig:body-Gazebo} shows the simulated RGB-D camera and human body in the Gazebo, and the synthetic point cloud produced from the simulation.
% \\
% \textcolor{blue}{MENTION HERE WHAT IS THE DISTRIBUTION OF LABELS (how many of each categories do you have)}
%

\subsection{Obtaining a Real Dataset from the ResQbot sensor }

The final aim of this work is to test and implement the ResQNNet casualty detector using real sensor data from the ResQbot. Fig. 6 shows the real point-cloud data obtained from the ResQbot sensor. The point-cloud dataset that we collect for the experiments includes point clouds that contain a casualty (human body lying on the ground) and point-clouds that contain other objects that are not casualties (furniture, boxes, wall, stairs etc). 

\begin{figure}[t]
\centering
\includegraphics[width=3.1in]{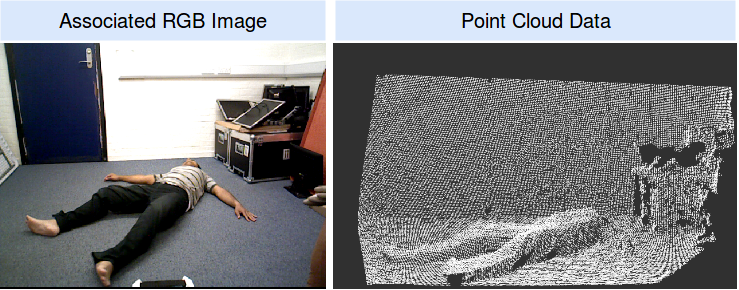}
\caption{Obtaining real point-cloud sensor data from ResQbot. \\
\textbf{Left}: Associated RGB image in which the point-cloud data is taken. \textbf{Right}: Visualisation of real point-cloud data from the RGB-D sensor.}
\label{fig:real-body}
\vspace{-0.4cm}
\end{figure}

\subsection{Augmenting the Dataset}
One of our hypotheses is that introducing additional specific augmentation strategies to the synthetic training data can increase the similarity between the simulation and reality.
%
% \textcolor{red}{In addition to this, similarly to domain randomisation approaches, training a model in such a way would make it robust to out-of-distribution samples which are different from simulated data.}
%
Hence, we expect that by incorporating these augmenting strategies with the training data, the trained model would also perform well with real sensor data inputs.  

%
% \textcolor{blue}{I DONT UNDERSTAND THIS PHRASE:\\
In addition to the standard data augmentation---rotation, scaling and shifting---performed on the synthetic GPPC heightmap (or image), % or we can refer to it as an image in this work, 
we also observe the effect of introducing additional augmenting strategies on the detector’s performance. These special augmentation strategies include: 

\begin{enumerate}
\item Introducing various noise to the GPPC image (i.e. direct input of the neural network), including Gaussian noise, salt-and-pepper noise (SnP) and periodic noise;  
\item Introducing noise to the raw point-cloud sensor data readings (i.e. the input of ResQNNet); 
\item Reducing the resolution of the point-cloud data (i.e. down-sampling the numbers of points); 
\item Partially removing segments of the point-cloud data (simulating partial observability or occlusions). 
\end{enumerate}
 
\begin{figure}[t]
\centering
    \begin{subfigure}[c]{0.5\textwidth}
        \includegraphics[width=3.4in]{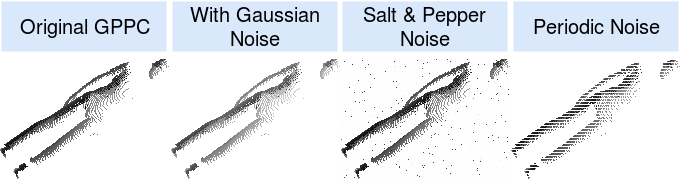}
        \subcaption{Adding noise into GPPC heightmap}
        \vspace{8px}
    \end{subfigure}
    ~~~~
    \begin{subfigure}[c]{0.5\textwidth}
        \includegraphics[width=3.4in]{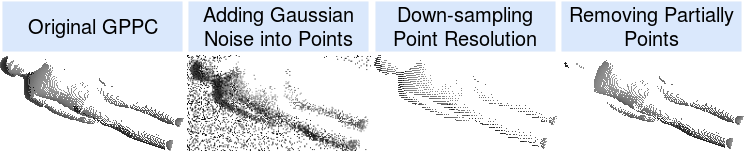}
        \subcaption{Adding noise into raw point clouds}
    \end{subfigure}    
\caption{Illustration of various data augmentations strategies used within the proposed framework.}
\label{fig:augmentation}
\vspace{-1.0em}
\end{figure}

Fig.~\ref{fig:augmentation} illustrates the effects of various of augmentation strategies to the GPPC image (i.e. heightmap) as well as to the raw point cloud data.

%DONE
%[NOT JUST TO THE GPPC BUT ALSO TO RAW PC]

% \subsection{Transfer Learning}
% In comparison to our propose sim-to-real approach, we also observe the transfer learning technique. 
% %
% Using this transfer learning technique, we utilize the feature extraction part from CNN network trained using the synthetic data.
% %
% The transfer learning strategy used for this study is outlined as following steps:

% \begin{enumerate}
% \item Firstly, we train the networks using synthetic data until achive the desired performance.
% \item After that, we freeze the the convolution parts of the network (i.e. conv 1, pool 1, conv 2, pool2, flatten), and remove the last fully-connected layers. 
% \item Then, we retrain the network using small dataset from real sensor data by treating the freezing layer as a fixed feature extractor and train the weights of the rest fully-connected layers.
% \end{enumerate}

\section{EXPERIMENTAL SETUP AND RESULTS}

%\begin{table}[t]
%\begin{center}
%  \caption{The combination of value settings for generating the synthetic %bodies.}
%  \label{tab:body-parameter}
%  \begin{tabular}{rc}
%    \toprule
%    Parameter&Value Range\\
%    \midrule
%    Stature (cm) &1500, 1600, 1700, 1800, 1900, 2000 \\
%    BMI (kg/m^2)&20, 25, 30 \\
%    SHS (ratio)&0.4, 0.5, 0.6 \\
%    Age&20, 40, 60, 80 \\
%    \bottomrule
%  \end{tabular}
%\end{center}
%\end{table}

\begin{figure*}[!t]
\centering
\includegraphics[width=6.9in]{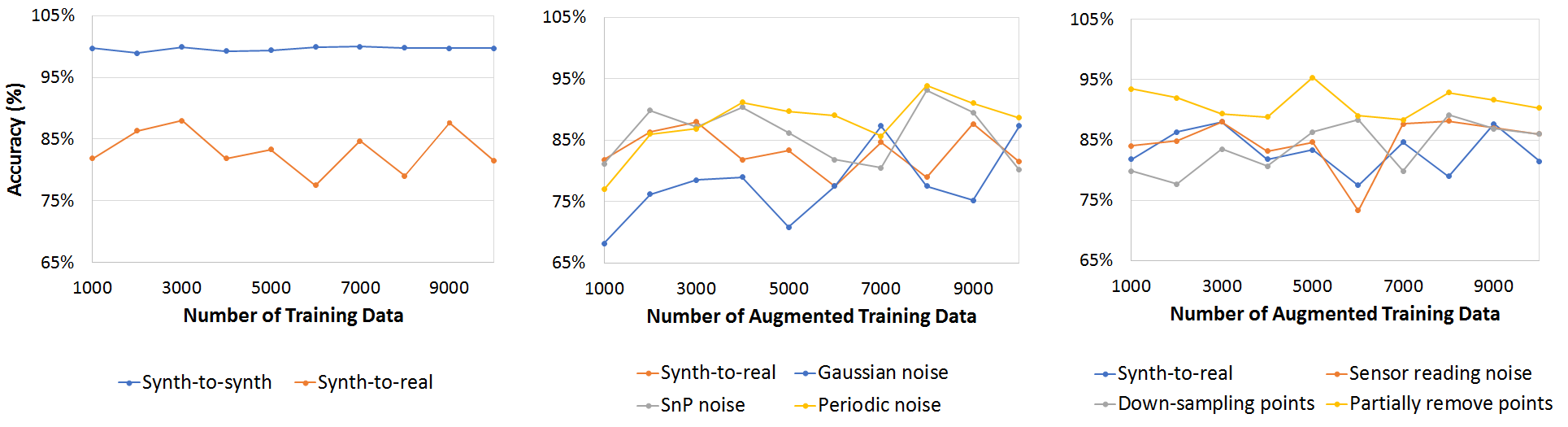}
\vspace{-0.2cm}
\caption{Effects of different augmentation strategies on the classification accuracy of models trained on synthetic data; \textbf{Left}) comparison of the performance on the synthetic and real test datasets, of models trained without data augmentation; Effects of different augmentation strategies: \textbf{Centre}) applied to GPPC, \textbf{Right}) applied directly on point-cloud data.}
\label{fig:results}
\vspace{-0.6cm}
\end{figure*}

% \begin{figure*}[!htb]
% \centering
% \includegraphics[align=t,height=4.40cm,trim={0.7cm 0 0.1cm 0},clip]{results_a.png}
% \includegraphics[align=t,height=4.80cm,trim={1.3cm 0 0 0},clip]{results_b.png}
% \includegraphics[align=t,height=4.80cm,trim={0 0 0.1cm 0},clip]{results_c.jpg}
% \caption{Effects of different augmentation strategies on the classification accuracy of models trained on synthetic data; \textbf{Left}) comparison of the performance on the synthetic and real test datasets, of models trained without data augmentation; Effects of different augmentation strategies: \textbf{Centre}) applied to GPPC, \textbf{Right}) applied directly on point-cloud data.}
% \label{fig:results}
% \vspace{-0.4cm}
% \end{figure*}

To quantitatively evaluate the performance of our proposed approach, we conducted several experiments and performed the ablation study to analyse the contribution of each of the data augmentation strategies and the number of corresponding augmented data samples.
% evaluation independently in several dataset consisting training, validation and testing datasets, as well as including synthetic and real sensor datasets. 
%
%
The network implementation and training in our experiments was done using Keras. 
The hyperparameters we used in the experiments were adopted from LeNet implementation for image classification~\cite{LeCun2006PROCOT}.

We trained the classification part of ResQNNet (the CNN adopted from LeNet) on all cropped images, i.e. candidate regions of the synthetic GPPCs, classified as casualty or non-casualty. 
For the purpose of performance analysis, we independently trained multiple CNN instances, one for each data augmentation strategy and the number of augmented samples it generated (incrementally increasing from 1000 to 10000 samples, with a step of 1000), and then evaluated the trained model. Parameter settings for each augmentation strategy can be found in Table~\ref{tab-param-setting}.

\begin{table}[]
\centering
\caption{The summary of the parameter settings used for each of the data augmentation strategy.}
\begin{tabular}{lll}
\hline
 & \textbf{Augmentation Type} & \textbf{Parameters \& Values} \\ \hline
\textbf{GPPC Images} & Gaussian noise & \begin{tabular}[c]{@{}l@{}}$\mu$: 0.2 \\ $\sigma$: 10, 25, 50\end{tabular} \\ \cline{2-3} 
 & Salt \& pepper noise & \begin{tabular}[c]{@{}l@{}}s-p ratio: 0.5\\ noise amount: 0.01-0.05\end{tabular} \\ \cline{2-3} 
 & Periodic noise & \begin{tabular}[c]{@{}l@{}}set row value to 255\\ - every 3, 5, 10 rows\end{tabular} \\ \hline
\textbf{Point Cloud Data} & Sensor noise & \begin{tabular}[c]{@{}l@{}}$\mu$: 0  \\ $\sigma$: 0.1, 0.2, 0.25\end{tabular} \\ \cline{2-3} 
 & Down-sampling & scale: 1/50, 1/20, 1/10 \\ \cline{2-3} 
 & Removing segments & size: 50, 100, 150 \\ \hline
\end{tabular}
\label{tab-param-setting}
\vspace{-0.4cm}
\end{table}

The training data is randomly shuffled and the class labels are equally distributed.
The evaluation is done on the leave-one-out synthetic test set to analyse the trained model's accuracy, as well as on the real point-cloud data obtained from the ResQbot's RGB-D sensor to evaluate the model's generalisation capabilities.
It is important to emphasise that the real sensor data is not used during any of the training instances, but only for evaluation purposes.
%
%large synthetic dataset in hierarchical level. 
%
%Each scenario that will be evaluated—including each data augmenting strategy is also pre-trained separately. 
%
To quantify the performance of the trained model on both the synthetic test data and real sensor test data, we calculate the classification accuracy as the performance measure, as well as the $F_1$ score, summarised in the Table~\ref{tab:results}.
Fig.~\ref{fig:results} shows the classification accuracy, for each of the instances mentioned above, as a function of the number of augmented data samples.

\textbf{Performance on the synthetic test data}: We evaluate the trained models on 1000 previously unseen synthetic data samples. The results shown in Fig.~\ref{fig:results}a demonstrate that most of the trained models result in accurate predictions on the synthetic test data with up to 99\% accuracy (see Table~\ref{tab:results}).

\textbf{Performance on the real sensor test data}: 
We further evaluate the trained models on the real sensor data obtained from the ResQbot's sensor reading. This dataset is split into two groups: validation data consisting of 300 samples of point-cloud data containing a casualty and 300 samples that do not contain a casualty; and testing data which consists of 100 samples of point-cloud data containing a casualty and 100 samples that do not contain a casualty. Fig.~\ref{fig:results}a, shows that the prediction accuracy of the trained models significantly drops (down to 80\%) when classifying real sensor data inputs (see Table~\ref{tab:results}).

\textbf{The effect of GPPC image data augmentation}: 
We hypothesised that augmenting synthetic training data could help improve the performance of the trained models in classifying real sensor data. To show this, we first performed data augmentation strategies to the GPPC images as direct inputs of the classifier network. We injected Gaussian, salt and pepper, as well as periodic noise to the image. The noise settings used in this experiments are described as follows:
\begin{enumerate}
    \item \textit{Injected Gaussian noise}: 
    In this experiments we injected random Gaussian noise to each pixel of the GPPC images. We used random noises with mean 0.2 and variance 10-50.
    % \textcolor{blue}{WHY IS MEAN NOT ZERO?}
    \item \textit{Salt and pepper noise}: 
    We introduced a combination between salt (i.e. set pixel value to max.) and  pepper (i.e. set pixel value to min.) in random pixels with a constant salt and pepper ratio (1:1) and with the noise contribution of 0.01-0.05.
    \item \textit{Periodic noise}: 
    For introducing periodic noise to the GPPC images, we set the value of all pixels in one row to 255 (i.e. max value), and we repeat it every 3, 5 and 10 rows.
\end{enumerate}

The results demonstrated in Fig.~\ref{fig:results}b and Table~\ref{tab:results} indicate that in these experiments, the noise introduced to the GPPC images have only a small impact on improving the performance of the trained model classification of real sensor data.
Salt and pepper (SnP) noise introduced into the GPPC training images bring the highest improvement to the accuracy (up to 84\%).
However, in our conducted experiments, Gaussian noise injected in the images remarkably reduces the performance of the networks. 
One reason for this might be because Gaussian noise is not a good approximation of the noise occurring on the real GPPC images.

\textbf{The effect of raw point-cloud data augmentation}: Since we proposed to use point cloud data as inputs, we also hypothesised that augmenting the raw point cloud data inputs could directly affect the performance of the trained models. To simulate imperfections of real point cloud data produced by the RGB-D sensor and the disturbances from environmental conditions, we experimented with adding Gaussian noise to the point clouds, down-sampling the points and partially removing point segments. The augmentation settings applied to the point cloud data used in this experiments are the following:
\begin{enumerate}
    \item \textit{Simulated sensor noise}: 
    In real applications, the sensor measurements are not perfect and they contain measurement noise. The most common noise occurring in sensor measurements is white noise, which can be modelled as Gaussian noise with zero mean and a certain variance $\mathcal{N}(0, \sigma)$. To model this noise, we use $\sigma=0.1, 0.2, 0.25$ based on the experiment results reported in~\cite{mirdanies2017experimental}.
    \item \textit{Down-sampling point cloud data}: 
    We introduce the down sampling augmentation strategy applied to the point cloud data, in order to model different sensor resolutions. In our experiments, we introduced down-sampling to the original point cloud data with a scale 1:50, 1:20 and 1:10 to augment the data.
    \item \textit{Partially removing point segments}: 
    In real life the point cloud data obtained from RGB-D sensor often returns NaN values when detecting a surface with bad reflective properties, or in bad lighting conditions. 
    To model this imperfection, we introduce an augmentation strategy that partially removes random rectangular segments of the synthetic point cloud data used for the training process. 
\end{enumerate}

The results demonstrated in Fig.~\ref{fig:results}c show that in contrast with the previous augmenting strategies that target the GPPC images, these strategies significantly affect the performance of the trained models. 
Specifically, the results of the models trained on data augmented by partially removing point-cloud segments indicate that this strategy could considerably improve the model's performance to up to 91\% accuracy when applied to real sensor data (see Table~\ref{tab:results}).

\begin{table}[t]
\begin{center}
  \caption{Summary of the experimental results, which shows the average performance of each data augmentation strategy used in the experiments.}
  \label{tab:results}
  \begin{tabular}{rcccc}
    \toprule
    Training \& \\ Augmentation & &Accuracy  & &F1 Score\\
    \midrule
    Synth-to-synth& & 99.63 & & 96.78\\
    Synth-to-real& & 83.16 & & 79.85\\
    \midrule
    + Gaussian noise& & 77.75 & & 71.48\\
    + SNP noise& & 84.68 & & 82.38\\
    + Periodic noise& & 83.81 & & 80.93\\
    \midrule
    + Sensor reading noise& & 85.98 & & 83.78\\
    + Down-sampling points& & 87.88 & & 86.28\\
    + Removing points& & 91.11 & & 90.65\\
    \bottomrule
  \end{tabular}
\end{center}
\vspace{-0.8cm}
\end{table}

\textbf{The effect of the augmentation combination strategy}: 
In addition to the independent augmentation experiments, we also investigated the effect of possible combination of each augmentation strategy to improve the sim-to-real performance.
We used the Bayesian Optimisation (BO) approach to find the optimal combination of the augmentation strategies that yields the highest model performance.
To find this, we used a validation set of real sensor data that includes 300 point cloud data with a casualty in the data and another 300 point cloud data with a non-casualty.

\begin{figure}[t!]
\centering
\includegraphics[width=3.4in,trim={0cm 0cm 1cm 0}]{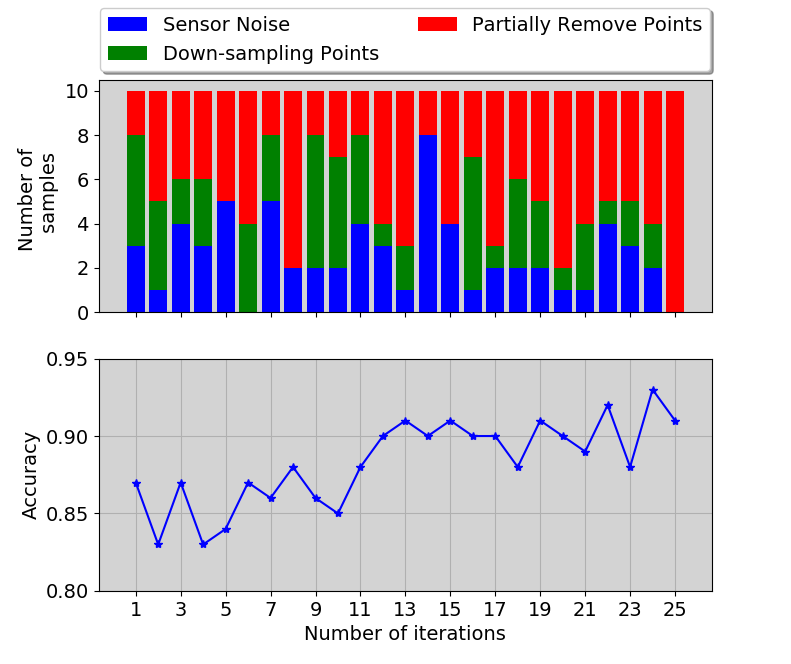}
\vspace{-0.5cm}
\caption{Plots showing the contributing sample proportions coming from each data augmentation approach (legend on top) and the corresponding accuracy achieved on the synthetic point cloud training data. The $y$ axes of the top and the bottom plots show the number of samples in thousand and the accuracy, respectively. The 25 iterations shown are obtained via Bayesian Optimisation~\cite{Fernando2018}.
}
\label{fig:bo-process}
\vspace{-0.5cm}
\end{figure}

\begin{figure*}[!htb]
\centering
\includegraphics[align=t,height=5.2cm,trim={1cm 0cm 5.6cm 0},clip]{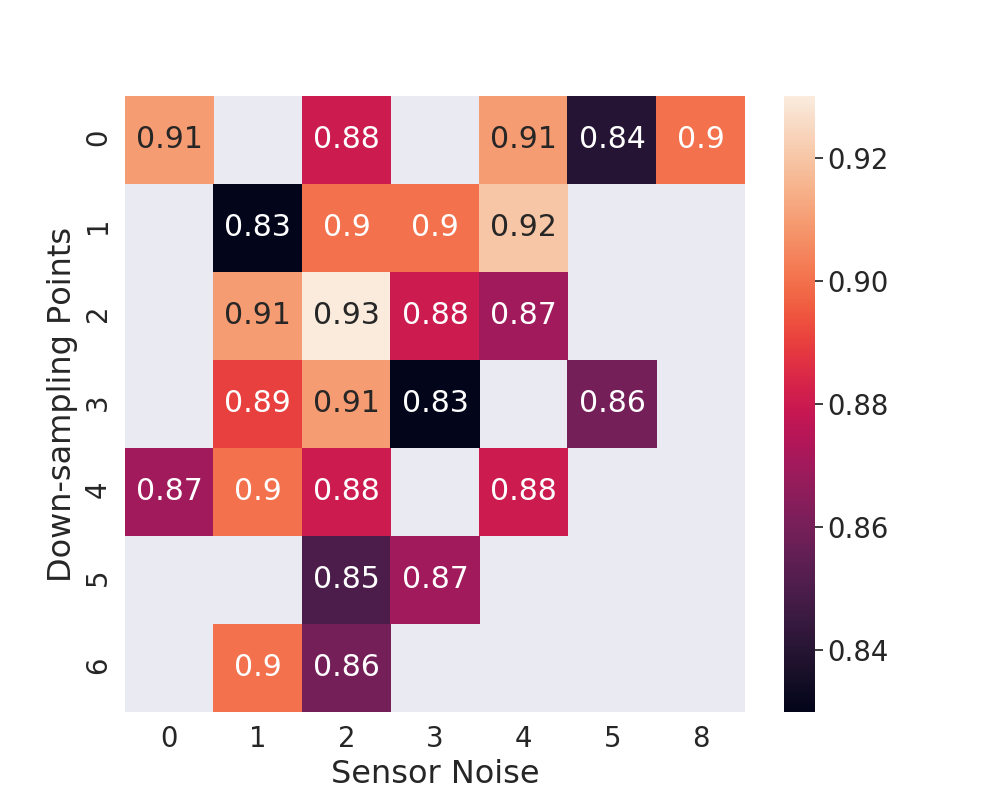}
\includegraphics[align=t,height=5.20cm,trim={0cm 0 5.6cm 0},clip]{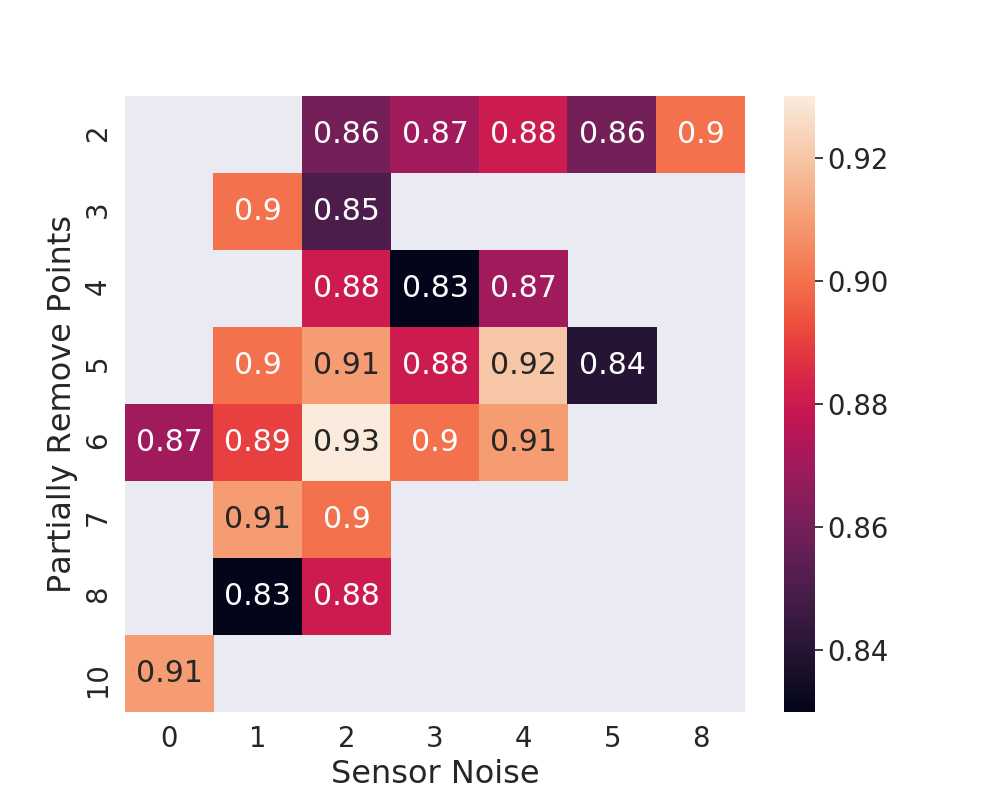}
\includegraphics[align=t,height=5.20cm,trim={0 0 2cm 0},clip]{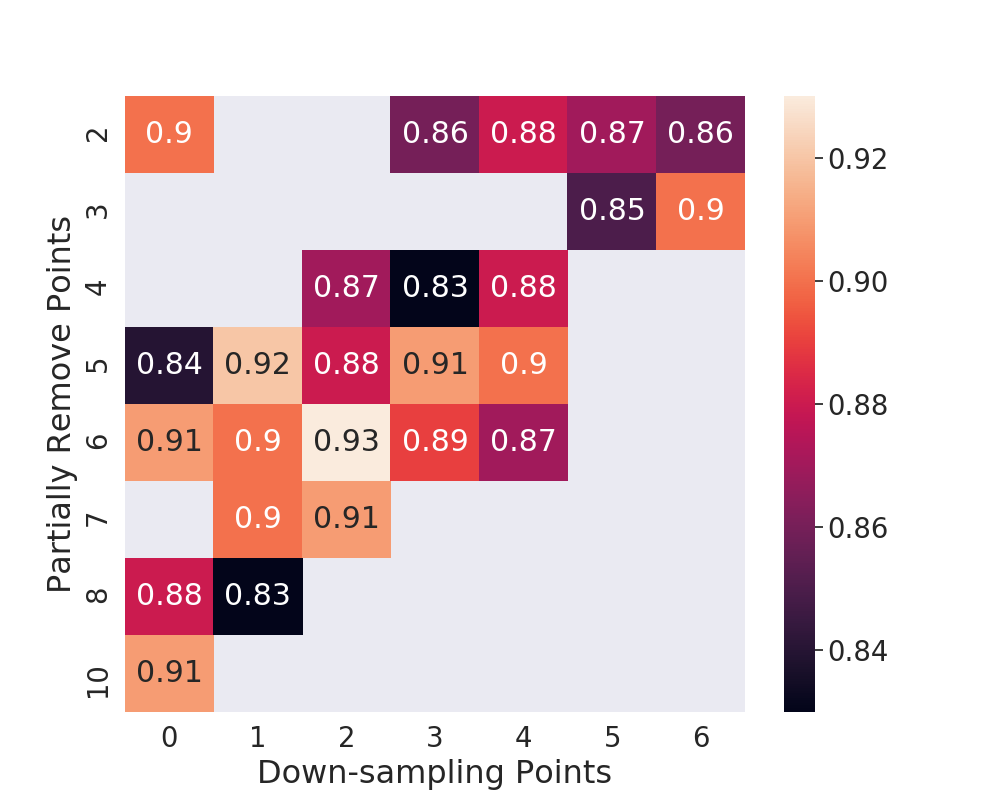}
\vspace{-0.1cm}
\caption{Heatmaps showing the accuracy achieved for different combinations of data augmentation techniques. 
Each plot is a projection on a 2-dimensional plane corresponding to a pair of data augmentation techniques.
The axes show the number of samples in thousand, and the color intensities represent the accuracy, as indicated in the colormap on the right.}
\label{fig-bo-results}
\vspace{-0.6cm}
\end{figure*}

In these experiments, we adapted the implementation from~\cite{Fernando2018}.
We used the default BO hyperparameters and performed the optimisation process in 25 iterations.
The result of this hyperparameter optimisation process can be found in Figs.~\ref{fig:bo-process}-\ref{fig-bo-results}.
We found that the optimal combination of the augmentation strategies consisted of: 1) 2000 samples augmented by adding sensor data noise, 2) 2000 samples augmented by down-sampling points, and 3) 6000 samples augmented by partially removing point segments.
The results show that this combination can improve the performance of sim-to-real learning in term of detection accuracy in real sensor data up to $93\%$ in validation data.
We then tested the trained network from this optimal combination using unseen testing data consisting 100 point cloud data with casualty in it and another 100 point cloud data with no-casualty in it.
The result shows that the trained network achieves $95\%$ accuracy on real sensor data test set, with a balance of false positive and false negative errors.

% \begin{figure*}[h!]
% \centering
% \includegraphics[width=6.7in]{heat_map.png}
% \vspace{-0.2cm}
% \caption{Heatmaps showing the accuracy achieved for different combinations of data augmentation techniques. 
% Each plot is a projection on a 2-dimensional plane corresponding to a pair of data augmentation techniques.
% The axes show the number of samples in thousands, and the color intensities represent the accuracy, as indicated in the colormap on the right.
% }
% \label{fig-bo-results}
% \vspace{-0.5cm}
% \end{figure*}

\section{CONCLUSIONS AND FUTURE WORK}
In this study, we have addressed the problem of casualty detection from point cloud data.
More specifically, we proposed a novel sim-to-real domain-randomisation-based learning technique in which we trained our model to perform casualty vs non-casualty classification from synthetic point-cloud data.
We introduced several data augmentation strategies for synthetic data, and observed the effect of each strategy, as well as their combinations, on the performance of the models trained with this data, when applied to the real sensor data test dataset. 
The experimental results demonstrate that the data augmentation strategies on raw point-cloud data
% ---including injecting noise, down-sampling point cloud resolution and removing, partially, the points---
have contributed to superior model testing performance when compared to the data augmentation done using GPPC heightmap images.
% ---produced from ground projected point .
%
In particular, augmenting synthetic point-cloud data by partially removing point segments, has shown considerable effect on improving the trained model's classification performance on the real sensor test data, compared to other strategies.
This can be explained as such a strategy simulates scenarios which occur in reality; e.g. the RGB-D sensor often gives NaN values when detecting a surface with bad reflective properties, or when bad lighting conditions also affect the quality of the point-cloud data from the RGB-D sensor. 
%
% Moreover, lighting condition is also affecting the quality of returned point cloud data from the RGB-D sensor.
%

As future work, we would like to extend this approach and examine possible implementations using point-cloud data produced by 3D LIDAR sensors, which have different properties than RGB-D sensors.
Moreover, we will investigate other factors that can be randomized and would contribute to the robustness of the estimator.
Some of the techniques to explore would be transfer learning and fine tuning to improve the final performance.
%

%%%%%%%%%%%%%%%%%%%%%%%%%%%%%%%%%%%%%%%%%%%%%%%%%%%%%%%%%%%%%%%%%%%%%%%%%%%%%%%%
\vspace{-0.1cm}
\section*{ACKNOWLEDGMENT}

Roni Permana Saputra would like to thank the Indonesia Endowment Fund for Education -- LPDP, for the financial support of the PhD program.  
The authors also would like to show our gratitude to James Paul Foster for sharing his comments and feedback on this work.

%%%%%%%%%%%%%%%%%%%%%%%%%%%%%%%%%%%%%%%%%%%%%%%%%%%%%%%%%%%%%%%%%%%%%%%%%%%%%%%%

\bibliographystyle{IEEEtran}
\bibliography{main}

\end{document}